# Preference Incorporation into Many-Objective Optimization:

# An Outranking-based Ant Colony Algorithm

Gilberto RIVERA[1], Carlos A. COELLO COELLO[2], Laura CRUZ-REYES[3], Eduardo R. FERNANDEZ[4], Claudia GOMEZ-SANTILLAN[3]*, and Nelson RANGEL-VALDEZ[3]

[1] División Multidisciplinaria de Ciudad Universitaria, Universidad Autónoma de Ciudad Juárez, 32579, Cd. Juárez, CHI, Mexico

[2] CINVESTAV-IPN, Evolutionary Computation Group, 07300, CDMX, Mexico

[3] Postgraduate and Research Division, National Mexican Institute of Technology/Madero Institute of Technology, 89440 Ciudad Madero, TAMPS, Mexico

[4] Universidad Autónoma de Coahuila, Facultad de Contaduría y Administración, Blvd. Revolución Oriente No. 151, 27000, Torreón, México.

* Corresponding autor

**Abstract**. In this paper, we enriched Ant Colony Optimization (ACO) with interval outranking to develop a novel multiobjective ACO optimizer to approach problems with many objective functions. This proposal is suitable if the preferences of the Decision Maker (DM) can be modeled through outranking relations. The introduced algorithm (named Interval Outranking-based ACO, IO-ACO) is the first ant-colony optimizer that embeds an outranking model to bear vagueness and ill-definition of DM preferences. This capacity is the most differentiating feature of IO-ACO because this issue is highly relevant in practice. IO-ACO biases the search towards the Region of Interest (RoI), the privileged zone of the Pareto frontier containing the solutions that better match the DM preferences. Two widely studied benchmarks were utilized to measure the efficiency of IO-ACO, i.e., the DTLZ and WFG test suites. Accordingly, IO-ACO was compared with two competitive many-objective optimizers: The Indicator-based Many-Objective ACO and the Multiobjective Evolutionary Algorithm Based on Decomposition. The numerical results show that IO-ACO approximates the Region of Interest (RoI) better than the leading metaheuristics based on approximating the Pareto frontier alone.

# 1 Introduction

Many engineering, science, and industry problems require considering the simultaneous optimization of several conflicting objective functions. Multi-Objective Evolutionary Algorithms (MOEAs) have been applied successfully in solving problems with 2-3 objective functions, but in real life, they often involve more objectives. Most MOEAs have limitations when trying to adequately solve problems with four or more objectives (the so-called Many-objective optimization problems, MaOPs). Bechikh et al. (2017) summarized the challenges faced by the state-of-the-art MOEAs in solving MaOPs:

1) Many solutions in the current MOEA's population become non-dominated, weakening the selective pressure toward the true Pareto front.

2) The increasing number of dominance-resistant solutions makes it difficult to discriminate well enough among solutions; these are solutions with poor values in some objectives but with near-optimal values in others (Ishibuchi et al., 2020).
3) Reduced effectiveness of the genetic operators (crossover, mutation, and selection)
4) Difficulty in representing the known Pareto front, since (given a resolution level) the number of points grows exponentially;
5) Complex solutions visualization in a high-dimensional space;
6) The higher computational cost derived from the estimation of diversity measures.

The above difficulties are severe in Pareto dominance-based MOEAs. However, some challenges (particularly Points 1, 3, and 6) are less demanding in decomposition-based algorithms, like MOEA/D (Zhang & Li, 2007), or non-evolutionary metaheuristics that build independent solutions. Nevertheless, identifying an approximation to the Pareto front is not enough; the decision-maker (DM) requires finding the best compromise solution; this solution is the one "most in agreement" with tthe DM's preferences. The completion of this process requires articulating the DM's preferences, which can be carried on at different stages of the decision-making process: a priori, 'a posteriori', or interactively.

In 'a posteriori' incorporation of preferences, the DM has to choose the final solution once the metaheuristic algorithm has generated the approximated Pareto frontier. An assumption behind this strategy is that the set of solutions must contain those that are the most satisfactory for the DM. This privileged set of solutions is called the Region of Interest (ROI). Another implicit assumption is the DM's capacity to identify his/her best solution, even in many-objective problems. The options to perform the selection process are the following ones:
  a) To make a heuristic selection. This difficulty of this task increases with the space dimensionality due to the cognitive limitations of the human mind (Miller, 1956).
  b) To use a formal multi-criteria decision method, which involves a model of the DM's preferences.

A priori and interactive preference incorporation approaches give relevant advantages over the a posteriori strategy:
  a) An increment of selective pressure toward solutions closer to the ROI, thus narrowing the search space (Branke, et al., 2016) and helping to find better solutions.
  b) As a consequence of Point a), a decrement in the number of candidate solutions to be the best compromise, which reduces the DM's cognitive effort to choose the final solution.

As a consequence of the above advantages, the interest in combining MOEAs and multicriteria decision-making (MCDM) techniques has increased in recent years. There are many proposals to incorporate preferences into the metaheuristic search processes that exist in the scientific literature. Regardless of the stage of the preferences' articulation, the following general strategies group a vast majority of the approaches:
- Expectation-based methods
- Comparison of objective functions
- Comparison of solutions
- Preference relations that replace dominance

The expectation strategy refers to goals that the DM wants to achieve for objectives (Xin et al., 2018); it contains those works that use reference points (e.g., Siegmund et al., 2017; Yutao et al., 2019; Liu et al., 2020; Li et al., 2020; Wang, et al., 2021; Abouhawwash and Deb, 2021), and desirability thresholds (e.g., Wagner, 2010; He et al., 2020). The comparison of objective functions includes those works that use weights for the objective functions (e.g., Branke and Deb, 2005; Brockhoff et al., 2013; Wang et al., 2015), ranking of objective functions (e.g., Cvetkovic and Parmee, 2002; Taboada et al., 2007; Kulturel-Konak et al., 2008), and trade-offs between objective functions (e.g., Branke et al., 2001; Miettinen et al., 2008). The comparison of solutions has different methods such as the ranking of solutions (e.g., Deb et al., 2010; Yuan et al., 2021), pairwise comparisons (e.g., Branke et al., 2016; Tomczyk and Kadzinski, 2020), classification of solutions (e.g., Cruz-Reyes et al., 2017; Cruz-Reyes et al., 2020), scoring (e.g., Li, 2019; Saldanha et al., 2019), and characterization of the preferred region (e.g., Dunwei et al., 2017). The last group refers to those methods that use preference relations instead of Pareto dominance (e.g. Parreiras and Vasconcelos, 2007; Fernández et al., 2010; Fernández et al., 2011; Helson et al., 2018; Balderas et al. 2019; Yi et al., 2019).

Many of the above methods should be used interactively (e.g., those based on the ranking of solutions, pairwise comparison of solutions, and classification of solutions). Other methods can be used in an a priori and progressive way (e.g., those based on preference relations that replace dominance). Interactive methods are trendy due to their following features: (i) the algorithm capacity to 'learn' the DM's preferences, thus suggesting more preferred solutions (Tomczyk and Kadzinski, 2020); (ii) the DM learns about her/his problem and updates her/his multi-criteria preferences; and (iii) since the DM is involved in the search process, (s)he feels more comfortable with the solutions found. However, the main criticism of interactive methods is because they require transitive and comparable preferences from the DM (French, 1993). Problems with a few objective functions fulfill these requirements. But when the dimension increases, these demands can become very hard for most real DMs.

In contrast, an a priori incorporation of preferences does not demand a rational behavior from the DM. Still, it requires a direct preference parameter elicitation that is only viable with significant imprecision. To alleviate this drawback, Fernandez et al. (2019) proposed an interval-based outranking method, which can handle non-transitive preferences incomparability and veto situations, such as the ELECTRE multi-criteria decision methods (e.g., Roy, 1990). These properties are relevant for solving real-world problems because the preferences of many DMs are non-compensatory and non-transitive. The interval outranking method of Fernandez et al. (2019) allows incorporating imprecisions in preference parameter values when setting 'a priori.' The DM feels more comfortable eliciting the preference parameter values as interval numbers than as precise values; this is even more important when the DM's preferences are ill-defined (e.g., the DM is a collective entity), or when the DM is an inaccessible person (e.g., the CEO of a very important enterprise). If the DM cannot provide a direct parameter, the indirect elicitation method of Fernández et al. (2020) allows inferring all the required parameters. A variant of the interval outranking from (Fernández et al., 2019) was used by Balderas et al. (2019), incorporating "a priori" preferences in an evolutionary algorithm.

As one can see in the long list of papers referred above, there is vast research on incorporating preferences in MOEAs; comparatively, there are few works combining preferences with other metaheuristics. One of the most popular non-evolutionary metaheuristics is ant colony optimization (ACO), which is the subject of this paper. The ACO algorithms are inspired by the collective search behavior of food of certain species of ants, where the ants communicate indirectly with other members of the ant colony. Communication is based on modifying the local environment by depositing a chemical called pheromone. In the foraging for food, some species use the behavior called "trace-trace" and "Follow-trace" to find the shortest route between the nest and the food source (Grassé, et al., 1959). The ACO algorithms take some important characteristics from actual ant colonies: a) indirect communication through traces of pheromones; b) shorter paths have higher pheromones; c) ants prefer paths with higher amounts of pheromone. Initially, ACO's design was for solving combinatorial optimization problems (Dorigo et al., 2004). For solving complex multiobjective NP-hard problems, the long-time optimization of multiobjective ant colonies has been applied as a powerful search technique. Consequently, ACO approaches attracted the attention of an increasing number of researchers, and many successful applications are now available (Mandal 2020, Chandra 2012, Dorigo 2005). Multiobjective Ant Colony Optimization (MOACO) algorithms have been used to solve discrete optimization problems in various domains.

ACO and MOACO algorithms generally concern specific problems, typically discrete problems. This is due to the ant colonies' characteristic to use specific information of the problem faced; these algorithms do not usually address general problems. The work of Falcón-Cardona and Coello Coello (2016) has the merit of proposing a way to exploit ant colonies without the need for specific knowledge of the problem but from the evaluation of the objective functions, which makes this approach independent of the problem and gives it greater generality. Falcon-Cardona and Coello Coello (2017) presented an extended version of the preliminary work reported in Falcón-Cardona and Coello Coello (2016), in which the indicator-based multiobjective ant colony optimization algorithm for continuous search spaces ($iMOACO_R$) was introduced. To the best of the authors' knowledge, this is the single MOACO algorithm explicitly designed to solve many-objective continuous optimization problems; it exhibits a competitive performance compared with state-of-the-art MOEAs.

In our view, there is nothing to prevent the use of preference incorporation strategies in MOACO algorithms; however, there are very few works within such an avenue of research. To contribute to close this research gap, this paper addresses the incorporation of DM preferences in a MOACO by using the interval outranking approach by Fernández et al. (2019). The proposed MOACO uses a preference relation built from the interval outranking information, combined with Falcón-Cardona and Coello Coello's (2017) strategy to solve many-objective continuous optimization problems. To our knowledge, no previous papers have incorporated preferences using this approach. Besides, the outranking approach is advantageous because it confers desirable properties to handle non-compensatory preferences and veto effects. The use of intervals allows handling imprecision and uncertainty in model parameters specified directly, also dealing with poorly defined preferences.

The organization of the remainder of this paper is as follows: Section 2 presents a review of related works to multiobjective optimization with ACO algorithms. Section 3 provides some background information, including a short description of the interval outranking method of Balderas et al. (2019), which slightly simplifies the proposal by Fernandez et al. (2019). Section 4 details the proposed Interval Outranking-based Ant Colony Optimization (IO-ACO) and the way used to incorporate the DM's preferences. Section 5 describes the experimental design, including results, discussion, and partial conclusions. Finally, Section 6 provides some general conclusions and future work.

## 2 Related Work of Preference Incorporation in Multiobjective Ant Colony Algorithms

ACO is a constructive method that forms solutions step by step by adding a solution component to the current partial solution. The DM's preferences can be incorporated into the construction of solutions since the transition rule, also called transition probabilities, can be easily redefined with preferences in mind. Nevertheless, to our knowledge, only a few works have incorporated preferences in MOACO algorithms.

Du et al. (2011) proposed a multiobjective scheduling problem in a hybrid flow shop. Two objectives considered in the proposed model are to minimize makespan and energy consumption. These two objectives are often in conflict. The Preference Vector Ant Colony System (PVACS) allows focusing the search in the solution space on particular decision-maker interest areas instead of searching for the entire Pareto frontier. This is achieved by maintaining a separate pheromone matrix for each objective and assigning a preferences vector to the ants. This vector, provided by the users, represents the relative importance of different objectives.

Chica et al. (2010, 2011) introduced some procedures for incorporating preference information into a MOACO algorithm called Multiple Ant Colony System (MACS). These procedures use an a priori approach to incorporate the Nissan managers' expertise eliciting preferences in both the decision variable and the objective space. In decision space, the procedure concerns only solutions having the same objective values. A discrimination procedure uses new relations formulated through preference measures, which consider expert-relevant requirements concerning decision variables. In objective space, the preferences are incorporated through two alternative approaches: (a) by units of importance and (b) by setting a set of goals. In (a), experts set units of importance to the achievement of the objectives. The definition of dominance is modified to specify acceptable trade-offs for each pair of objectives in terms of importance units. New objectives replace the original ones in an aggregation of them using the units of importance. For the second alternative, experts define goals and incorporated them into the objective function using well-known techniques to transform goals into MOO problems (Chica et al., 2009).

Cruz et al. (2014) optimized interdependent projects portfolios. They adapted an ACO metaheuristic to incorporate preferences based on the outranking model by Fernandez et al. (2011); this model reduces problems with several or many-objective functions to a surrogate three-objective optimization problem, which is solved through a lexicographic approach. The proposed non-outranked ACO (NO-ACO) algorithm searches for optimal portfolios in synergetic conditions and can handle interactions impacting both objectives and costs. Redundancy is also considered during portfolio construction. Since the selective pressure toward a privileged zone of the Pareto frontier increases by incorporating preferences, a zone that better matches the DM's preferences can be identified. Compared with other metaheuristic approaches that do not incorporate preferences, NO-ACO achieves greater closeness to the region of interest with less computational effort.

Developing the idea from Cruz et al. (2014), Fernandez et al. (2015) proposed the Non-Outranked Ant Colony Optimisation II method combined with integer linear programming (ILP) for optimizing project portfolio problems (PPP), with interacting projects and decisions of partial support to candidate projects. The advantages of this approach are evidenced by a wide set of computer experiments on realistic size project portfolio selection problems.

Do Nascimento Ferreira et al. (2016) proposed an interactive approach to solving the Next Release Problem (NRP), employing an ACO algorithm. The purpose is to reach solutions that incorporate subjective aspects while optimizing other important metrics related to the engineering requirements. The algorithm interacts with the user, showing all of her/his possible software requirements. (S)he selects one expectation about the next software release for each requirement: whether it should be present, or not, and no preference. This preference information is aggregated in a weighted sum single-objective function, aiming to maximize customer satisfaction. The last measure is calculated as the number of expectations met by a solution divided by the total number of expectations declared by the user. After that, the preference information is used to build solutions as part of the heuristic information of ACO.

Fernandez et al. (2017) presented an extension of the study of Bastiani et al. (2015). They solved a project portfolio selection problem with scarce information about the candidate projects. The strategy balances the priorities and the number of projects in the final portfolio through a model that minimizes discrepancies between the rank order of projects and the portfolio and maximizes its cardinality. Since the model considers many objective functions, the authors reduce the original many-objective problem to a surrogate three-objective functions. The surrogate problem is solved by an ACO algorithm like the one proposed by Cruz et al. (2014).

## 3. Background

This section presents the foundations of IO-ACO, our optimizer. Subsection 3.1 briefly describes ACO$_R$, the baseline ACO algorithm for continuous domains; Section 3.2 presents the interval-outranking approach for identifying the best-compromise solution, which follows the perspective of the European School of Multicriteria Decision Analysis.

*3.1 Ant Colony Optimization for Continuous Domains*

Ant Colony-based algorithms were initially designed to approach discrete optimization with a focus on graph problems. In general, ACO attempts to solve an optimization problem by iteratively:
  (a) Building solutions in a probabilistic way; here, the pheromone trail represents the probability distribution used.
  (b) Updating the pheromone trail according to the patterns in the best-evaluated solutions; this strategy is used to bias the next-iteration solutions towards high-quality regions in the search space.

A pivotal element for ACO algorithms is the pheromone representation, typically given in a numerical matrix denoted by $\tau$. The pheromone values act as a reinforcement learning model of the search experience of the ants.

ACO$_R$ (Socha & Dorigo, 2008) is probably the most remarkable ACO version available to optimize single-objective problems with continuous decision variables. Socha and Dorigo (2008) represented $\tau$ by an archive storing the best-so-far solutions. Given a problem with $m$ decision variables, a vector $x_l = \langle x_{l,1}, x_{l,2}, x_{l,3}, \ldots, x_{l,m} \rangle$ represents a solution, and $f(x_l)$ represents the objective function to minimize. Then, $\tau$ stores the $\kappa$ best-evaluated solutions, which are ascending sorted according to $f(x_l)$; the structure of $\tau$ is presented in Figure 1.

| | $x_{1,1}$ | $x_{1,2}$ | ... | $x_{1,j}$ | ... | $x_{1,m}$ | $f(x_1)$ | $\omega_1$ |
|---|---|---|---|---|---|---|---|---|
| $x_1$ | | | | | | | | |
| $x_2$ | $x_{2,1}$ | $x_{2,2}$ | ... | $x_{2,j}$ | ... | $x_{2,m}$ | $f(x_2)$ | $\omega_2$ |
| ⋮ | ⋮ | ⋮ | ⋱ | ⋮ | ⋱ | ⋮ | ⋮ | ⋮ |
| $x_l$ | $x_{l,1}$ | $x_{l,2}$ | ... | $x_{l,j}$ | ... | $x_{l,m}$ | $f(x_l)$ | $\omega_l$ |
| ⋮ | ⋮ | ⋮ | ⋱ | ⋮ | ⋱ | ⋮ | ⋮ | ⋮ |
| $x_\kappa$ | $x_{\kappa,1}$ | $x_{\kappa,2}$ | ... | $x_{\kappa,j}$ | ... | $x_{\kappa,m}$ | $f(x_\kappa)$ | $\omega_\kappa$ |
| | $G_1$ | $G_2$ | ... | $G_j$ | ... | $G_m$ | | |

Figure 1. Pheromone representation in ACO$_R$

According to Figure 1, each solution $x_l$ has an associated weight $\omega_l$, which measures the solution quality in terms of its position in $\tau$. The weight of the $l$th solution is defined as:

$$\omega_l = \frac{1}{\varsigma \cdot \kappa \sqrt{2\pi}} e^{-\varphi(l)}, \text{where } \varphi(l) = \frac{[l-1]^2}{2\varsigma^2 \kappa^2}, \quad (1)$$

which essentially defines the weight to be a value of the Gaussian function with argument $l$, mean 1.0, and standard deviation $\varsigma \cdot \kappa$, where $\varsigma$ is a parameter of the algorithm. When $\varsigma$ is small, the best-ranked solutions are strongly preferred, and when it is large, the probability becomes more uniform. The effect of this parameter on ACO$_R$ is to adjust the balance between the influences of the iteration-best and the best-so-far solutions.

According to Figure 1, there are $m$ Gaussian kernels $G_j$ $(1 \leq j \leq m)$, one for each decision variable. These kernels are used to infer a probability density function, expressed as

$$G_j(x) = \sum_{l=1}^{\kappa} \omega_l \, g_l^j(x).$$

A Gaussian kernel is a weighted sum of several one-dimensional Gaussian functions $g_l^j(x)$, defined as:

$$g_l^j(x) = \frac{1}{s_l^j \sqrt{2\pi}} e^{-\phi_j(l)}, \text{ where } \phi_j(l) = \frac{(x - x_{l,j})^2}{2(s_l^j)^2},$$

$g_l^j(x)$ defines a normal distribution where $x_{l,j}$ is the mean and $s_l^j$ is the standard deviation. The latter is dynamically calculated as ants construct solutions at each iteration.

An ant constructs a solution by performing $m$ construction steps. At the $j$th construction step, the $i$th ant infers a value for the variable $x_{i,j}$. As mentioned earlier, there are $m$ Gaussian kernels, each of them is composed of $\kappa$ regular Gaussian functions. However, at the $j$th construction step, only the resulting Gaussian kernel $G_j$ is required.

First, the weights $\omega_l$ are computed following Equation 1. Then, the sampling is performed in two phases. Phase one consists of choosing one of the Gaussian functions that compose the Gaussian kernel. The probability $p_l$ of choosing the $l$th Gaussian function is given by

$$p_l = \frac{\omega_l}{\sum_{r=1}^{\kappa} \omega_r}.$$

The choice of the $l$th Gaussian function is made once per ant and per iteration. This fact means that, during a complete iteration, the ants only use the Gaussian functions $g_l^j(x)$ associated with the chosen solution $x_l$ for constructing the solution incrementally in each step. This strategy allows exploiting the synergy among the decision variables.

Phase two consists of sampling the chosen Gaussian function ($l$). At the $j$th step, the standard deviation needs to be known for the single Gaussian function $g_l^j(x)$ chosen in phase one; consequently, only the standard deviation $s_l^j$ is needed. Indeed, the sampled Gaussian function differs at each $j$th construction step, and $s_l^j$ is dynamically calculated as follows

$$s_l^j = \xi \sum_{r}^{\kappa} \frac{|x_{r,j} - x_{l,j}|}{\kappa - 1},$$

which is the average distance from the chosen solution $x_l$ to other solutions in the archive and multiplied by the parameter $\xi$ $(1 \geq \xi > 0)$, which is the same for all the dimensions, having an effect like that of the pheromone evaporation rate in ACO. $\xi$ influences how the long-term memory is used, i.e., with high values, the search is less biased towards the points of the search space that have been already explored (cf. Socha & Blum, 2007). The $i$th ant randomly assigns the value of the $j$th decision variable following $x_{i,j} \sim g_l^j(x)$. The kernels $G_j(x)$ are probability density functions used to infer values of the decision variables for the near-optimal solutions and represent the reinforcement-learning model acquired by the colony.

Several researchers have recently extended ACO$_R$ to approach multiobjective optimization problems. Zhang et al. (2019) hybridized ACO$_R$ with support vector regression and chaos theory to solve the high-speed train nose problem; experiments with up to four objective functions were conducted in this case. Another main direction is to make changes into the pheromone matrix; for instance, MHACO (Acciarini et al., 2020) sorts the solutions of $\tau$ by hypervolume scores to solve space trajectory bi-objective test problems. iMOACO$_R$ (Falcon-Cardona and Coello Coello, 2017) performs this sorting considering the R2 metric (Brockhoff et al. 2012) to solve the DTLZ and WFG test suites with 3–10 objective functions. Our contribution is in line with this latter strategy: the Interval Outranking-based Ant Colony Optimization algorithms (IO-ACO) sorts the solutions following the ranking provided by interval outranking and preference relations to solve optimization problems with many-objective functions (3–10).

*3.2 The Relational System of Fuzzy Preferences based on Outranking*

The idea of incorporating the fuzzy outranking relations of ELECTRE into metaheuristics for many-objective optimization has been previously studied. In the domain of portfolio optimization, the pioneers of this strategy were Fernandez et al. (2010), who subsequently encouraged a wide range of studies in the last decade that exploits the properties of the outranking relations (e.g., Fernandez et al., 2011, 2013, 2017; Rivera et al., 2012; Cruz et al., 2014; Bastiani et al., 2015; Balderas et al., 2016; Gomez et al., 2018; Rangel-Valdez et al., 2020). These studies provide empirical evidence that metaheuristics increase selective pressure when enriched with DM preferences articulated through ELECTRE III. Consequently, they perform better than Pareto-based metaheuristics when many-objective optimization problems are faced.

The basis of the original idea is the relational system of preferences described by Roy and Vanderpooten (1996). A crucial model is $\sigma(x, y)$, which is the fuzzy value of the proposition '$x$ is at least as good as $y$' and calculated by classical methods from the literature (e.g., Roy 1990; Brans & Mareschal, 2005). The notion behind the fuzzy relational system of preferences proposed by Fernandez et al. (2011) is that solution $x$ is preferred over $y$ if '$x$ is at least as good as $y$' and '$y$ is not at least as good as $x$'—i.e., $\sigma(x, y)$ has a high value as $\sigma(y, x)$ has a low value.

ELECTRE defines $\sigma(x, y)$ considering
- the concordance index, $c(x, y)$, which measures the strength of the criteria coalition in favor of '$x$ is at least as good as $y$'; and
- the discordance index, $d(x, y)$, which measures the strength of the criteria invalidating the statement '$x$ is at least as good as $y$'.

On the one hand, in order to calculate the concordance index, it is necessary to know how the DM perceives the criteria and their values. This calculation requires the following parameters
- *Weight vector*: This represents how important each of the objectives is to the DM and is denoted by the vector $\vec{w} = \langle w_1, w_2, w_3, \ldots, w_n \rangle$, where $w_k > 0 \; \forall k \in \{1, 2, 3, \ldots, n\}$, $n$ is the number of objectives and $\sum_{k=1}^{n} w_k = 1$. Usually, the DM could hardly establish the value of each $w_k$, but they can utilize methods such as the revised Simos' procedure (Figueira & Roy, 2002) for this task.
- *Indifference thresholds*: This indicates how small the differences—in terms of objective values—should be for the DM to consider them as marginal or not significant on a practical level. Here, vector $\vec{q} =$

$\langle q_1, q_2, q_3, \ldots, q_n \rangle$ represents the indifference thresholds, where $q_k$ is the threshold for the $k$th criterion.

On the other hand, the discordance index is calculated based on the set of parameters known as the veto threshold. It is represented by the vector $\vec{v} = \langle v_1, v_2, v_3, \ldots, v_n \rangle$ and indicates the magnitude of the differences (in the objectives) between two alternatives to trigger a veto condition, being $v_k > q_k \forall k \in \{1,2,3,\ldots n\}$.

$c(x,y)$ is defined as the cumulative sum of the weights of the objectives for which $x$ is non-inferior to $y$ considering the indifference; the discordance index $d(x,y)$ introduces the following effect of rejection: if there is a difference against $x$ (according to the $k$th criterion) that exceeds $v_k$, then the predicate '$x$ is at least as good as $y$' is denied, regardless of the concordance index. $\sigma(x,y)$ combines both measures as $\sigma(x,y) = c(x,y) \cdot d(x,y)$.

Perhaps the strongest criticism of outranking models is the difficulty to find the precise values of the parameters (i.e., $\lambda$, $\vec{w}$, $\vec{q}$ and $\vec{v}$) which are unfamiliar for typical DMs, especially when the DM is a mythical entity (e.g., the public opinion), an inaccessible person, even an entity with ill-defined preferences and beliefs (e.g., a heterogeneous group). To mitigate this drawback, Fernandez et al. (2019) proposed an interval outranking, which can simultaneously handle multicriteria non-compensatory preferences and imperfect information in model parameters and criterion scores.

The rest of this subsection is structured as follows: Subsection 3.2.1 presents some basic notions on interval numbers, and Section 3.2.2 describes the relational system of preferences based on interval outranking.

*3.2.1 Preliminaries on interval numbers*

Interval numbers are an extension of real numbers and a subset of the real line $\mathbb{R}$ (cf. Moore, 1979). The representation of interval numbers throughout this document will be boldface italic letters, e.g., $\boldsymbol{E} = [\underline{E}, \overline{E}]$, where $\underline{E}$ and $\overline{E}$ correspond to the lower and upper limits. Among several operations of interval numbers, we are interested in the addition and the order relations defined below.

Let $\boldsymbol{D}$ and $\boldsymbol{E}$ be two interval numbers. The addition operation is defined as $\boldsymbol{D} + \boldsymbol{E} = [\underline{D} + \underline{E}, \overline{D} + \overline{E}]$. The relations $\geq$ and $>$ on interval numbers are defined by using the possibility function $P(\boldsymbol{E} \geq \boldsymbol{D})$. This function is defined as

$$P(\boldsymbol{E} \geq \boldsymbol{D}) = \begin{cases} 1 & \text{if } p_{ED} > 1, \\ p_{ED} & \text{if } 0 \leq p_{ED} \leq 1, \\ 0 & \text{otherwise}, \end{cases}$$

where $p_{ED} = \frac{\overline{E} - \underline{D}}{(\overline{E} - \underline{E}) + (\overline{D} - \underline{D})}$.

For the case when $\boldsymbol{D}$ and $\boldsymbol{E}$ are real numbers $D$ and $E$ (degenerate intervals), $P(\boldsymbol{E} \geq \boldsymbol{D}) = 1$ iff $E \geq D$; otherwise, $P(\boldsymbol{E} \geq \boldsymbol{D}) = 0$.

A realization is a real number $e$ that lies within an interval $\boldsymbol{E}$ (Fliedner and Liesio, 2016). Fernandez et al. (2019) interpret $P(\boldsymbol{E} \geq \boldsymbol{D}) = \alpha$ as the degree of credibility that once two realizations are given from $\boldsymbol{E}$ and $\boldsymbol{D}$, the realization $d$ will be smaller than or equal to the realization $e$. The relations $E \geq D$ and $E > D$ are defined by $P(E \geq D) \geq 0.5$ and $P(E \geq D) > 0.5$, respectively. These relations can also compare real numbers. The possibility functions meet the transitivity property because $P(E \geq D) = \alpha_1 \geq 0.5$ and $P(D \geq C) = \alpha_2 \geq 0.5 \Rightarrow P(E \geq C) \geq \min\{\alpha_1, \alpha_2\}$. Consequently, $\geq$ and $>$ are also transitive relations on interval numbers.

*3.2.2 The best compromise solution in terms of the interval outranking approach*

Below, we describe the interval outranking approach by Balderas et al. (2019), which slightly simplifies the proposal of Fernandez et al. (2019). First, the parameters of the outranking model are extended to become interval numbers or interval vectors. Thus, the weight vector is $\vec{w} = \langle w_1, w_2, w_3, \ldots, w_n \rangle$, where $w_k = [\underline{w_k}, \overline{w_k}]\ \forall k \in \{1,2,3,\ldots,n\}$, and $\sum_{k=1}^{n} \underline{w_k} \leq 1$ and $\sum_{k=1}^{n} \overline{w_k} \geq 1$. Similarly, the indifference-threshold vector is $\vec{q} = \langle q_1, q_2, q_3, \ldots, q_n \rangle$, where $q_k = [\underline{q_k}, \overline{q_k}]\ \forall k \in \{1,2,3,\ldots,n\}$ and the veto-threshold vector is $\vec{v} = \langle v_1, v_2, v_3, \ldots, v_n \rangle$, where $v_k = [\underline{v_k}, \overline{v_k}]$ and $v_k > q_k\ \forall k \in \{1,2,3,\ldots,n\}$. Finally, the majority threshold $\lambda = [\underline{\lambda}, \overline{\lambda}]$, where $\underline{\lambda} \geq 0.5$ and $\overline{\lambda} \leq 1$.

Let $x$ and $y$ two feasible solutions, and $f_k(x)$ and $f_k(y)$ their $k$th objective function. Let us consider the concordance coalition as the set $C_{x,y} = \{k \in \{1,2,3,\ldots n\} : P(-q_k \leq f_k(y) - f_k(x)) \geq 0.5\}$. The concordance coalition indicates the objectives favoring the statement '$x$ is at least as good as $y$'. A criterium that is not in the concordance coalition belongs to the discordance coalition $D_{x,y} = \{k \in \{1,2,3,\ldots n\} : k \notin C_{x,y}\}$.

Then, the concordance index $c(x,y) = [\underline{c(x,y)}, \overline{c(x,y)}]$ is a cumulative sum of the weights of the objectives belonging to $C_{x,y}$, whose limits are defined as

$$\underline{c(x,y)} = \begin{cases} \sum_{k \in C_{x,y}} \underline{w_k} & \text{if } \left(\sum_{k \in C_{x,y}} \underline{w_k} + \sum_{k \in D_{x,y}} \overline{w_k}\right) \geq 1, \\ 1 - \sum_{k \in D_{x,y}} \overline{w_k} & \text{otherwise,} \end{cases}$$

and

$$\overline{c(x,y)} = \begin{cases} \sum_{k \in C_{x,y}} \overline{w_k} & \text{if } \left(\sum_{k \in C_{x,y}} \overline{w_k} + \sum_{k \in D_{x,y}} \underline{w_k}\right) \leq 1, \\ 1 - \sum_{k \in D_{x,y}} \underline{w_k} & \text{otherwise.} \end{cases}$$

The discordance index is calculated as
$$d(x,y) = 1 - \max_{k \in D_{x,y}} \{P(f_k(x) - f_k(y) \geq v_k)\}.$$

Then, the outranking function is
$$\sigma(x, y) = \min\{P(c(x, y) \geq \lambda), d(x, y)\}. \quad (2)$$

Let $\beta$ be a threshold on the credibility of the statement '$x$ is at least as good as $y$' ($\beta \geq 0.5$). The binary relation S (outranking) is represented as
$$xSy = \{(x, y) : \sigma(x, y) \geq \beta\},$$
and the crisp relation '$x$ is preferred over $y$' is expressed as
$$x\text{Pr}y = \{(x, y) : x \preccurlyeq y \vee (xSy \wedge \neg ySx)\},$$
where the symbol '$\preccurlyeq$' stands for dominance in the Pareto sense.

From a set of feasible solutions $\mathcal{O}$, the following sets can be defined: $S(\mathcal{O}, x) = \{y \in \mathcal{O} : xSy\}$, and $P(\mathcal{O}, x) = \{y \in \mathcal{O} : y\text{Pr}x\}$. $S(\mathcal{O}, x)$ allows us to measure the strength of solution $x$, and $P(\mathcal{O}, x)$ allows us to measure its weakness. The best compromise $x^*$ from a set of solutions, $\mathcal{O}$ has the best values of strength and weakness and can be expressed as the bi-objective problem
$$x^* = \arg\min_{x \in \mathcal{O}}\{\langle |P(\mathcal{O}, x)|, -|S(\mathcal{O}, x)|\rangle\}, \quad (3)$$
with lexicographic priority in favor of $|P(\mathcal{O}, x)|$.

## 4. The Interval Outranking-based Ant Colony Optimization algorithm

IO-ACO is a many-objective optimizer that incorporates the DM's preferences to iMOACO$_R$, a remarkable version of ACO$_R$ for continuous problems with many objectives. According to Falcón-Cardona and Coello Coello (2017), iMOACO$_R$ has shown competitive results compared to algorithms that are a standard to approach many-objective optimization problems, i.e., MOEA/D and NSGA III. This reason was the primary motivation to propose an optimization algorithm extending iMOACO$_R$.

We changed iMOACO$_R$ to incorporate the interval outranking model. Initially, iMOACO$_R$ sorts the solutions in the pheromone structure $\tau$ according to their R2 scores (Brockhoff et al. 2012) —an indicator that measures uniformity on the distribution of solutions to establish different quality levels among Pareto-efficient solutions produced by the ants.

In IO-ACO, Pareto-efficient solutions are ranked by domination fronts which are obtained by considering the minimization of the two objectives $|P(\mathcal{O}, x)|$ and $-|S(\mathcal{O}, x)|$, according to the best-compromise definition given in Problem 3 ($|P(\mathcal{O}, x)|$ has lexicographic priority). Accordingly, the set composed by these fronts is denoted by $\mathcal{F} = \{\mathcal{F}_1, \mathcal{F}_2, \mathcal{F}_3, \ldots, \mathcal{F}_f\}$, where $\mathcal{F}_1$ contains the non-dominated solutions, $\mathcal{F}_2$ has the solutions that are dominated by only one solution, $\mathcal{F}_3$ those dominated by two solutions, and so forth. In general, $\mathcal{F}_r$ contains the solutions dominated by $r - 1$, where $1 \leq r \leq f$, and $f$ is the total number of levels. Having the solutions ranked, they are stored in $\tau$ together with their assigned rank. The solutions in $\mathcal{F}_1$ are the best-so-far vectors. Once $\tau$ contains the $\kappa$ best-ranked solutions, the mechanisms to generate new solutions can be applied as defined in ACO$_R$ (See Section 3.1).

Figure 2 represents the pheromone trail in IO-ACO. Here, there are $\kappa$ solutions with $m$ decision variables memorized in $\tau$. The solutions are ordered following the ranking given by $\mathcal{F}$, where, $\mathfrak{I}(x_l)$ is the front assigned to $x_l$; therefore, $\mathfrak{I}(x_1) \leq \mathfrak{I}(x_2) \leq \cdots \leq \mathfrak{I}(x_l) \leq \cdots \leq \mathfrak{I}(x_\kappa)$. Each weight $\omega_l$ is redefined as:

$$\omega_l = \frac{1}{\varsigma \cdot \Im(x_\kappa)\sqrt{2\pi}} e^{-\varphi(l)}, \text{ where } \varphi(l) = \frac{[\Im(x_l) - 1]^2}{2\varsigma^2 \Im(x_\kappa)^2}. \qquad (3)$$

Then, ACO$_R$ can be applied to construct solutions and approximate the best compromise solution. Algorithm 1 presents an outline of IO-ACO. First, $\kappa$ solutions are generated at random (Line 1) and normalized (Line 2) following the approach of Hernández Gómez and Coello Coello (2015) using $\alpha = 0.5$ and $\epsilon = 0.001$. Afterward, solutions in $\tau$ are ranked and sorted according to the domination front they belong to.

| $x_1$ | $x_{1,1}$ | $x_{1,2}$ | ... | $x_{1,j}$ | ... | $x_{1,m}$ | $\Im(x_1)$ | $\omega_1$ |
|---|---|---|---|---|---|---|---|---|
| $x_2$ | $x_{2,1}$ | $x_{2,2}$ | ... | $x_{2,j}$ | ... | $x_{2,m}$ | $\Im(x_2)$ | $\omega_2$ |
| $\vdots$ | $\vdots$ | $\vdots$ | $\ddots$ | $\vdots$ | $\ddots$ | $\vdots$ | $\vdots$ | $\vdots$ |
| $x_l$ | $x_{l,1}$ | $x_{l,2}$ | ... | $x_{l,j}$ | ... | $x_{l,m}$ | $\Im(x_l)$ | $\omega_l$ |
| $\vdots$ | $\vdots$ | $\vdots$ | $\ddots$ | $\vdots$ | $\ddots$ | $\vdots$ | $\vdots$ | $\vdots$ |
| $x_\kappa$ | $x_{\kappa,1}$ | $x_{\kappa,2}$ | ... | $x_{\kappa,j}$ | ... | $x_{\kappa,m}$ | $\Im(x_\kappa)$ | $\omega_\kappa$ |
|  | $G_1$ | $G_2$ | ... | $G_j$ | ... | $G_m$ |  |  |

Figure 2. Pheromone representation in IO-ACO. Gray elements are the same as in ACO$_R$, and black components are those adapted for IO-ACO

Lines 5–15 contain the main loop of the algorithm. The ants construct solutions (Lines 6–8) following the directions provided in Section 3.1. The previous solutions ($\tau$) and the new ones ($A$) are merged into $\mathcal{O}$ (Line 9). Then, the solutions in $\mathcal{O}$ are normalized, ranked, and sorted accordingly (Lines 10–11). The $\kappa$ solutions that better match the DM preferences are set in $\tau$ to influence the ants during the next iteration (Lines 12–14). When the algorithm finishes, IO-ACO provides the best compromise solution in $\tau$ (Line 16).

In terms of computational complexity, the most costly operation of Algorithm 1 is the assessment of the solutions to identify the best compromise (Line 11). The relation $xSy$ may be computed through $O(n)$ operations (see Equation 2), and the calculation of both objective functions of Problem 3 requires determining this relation for all pairs of solutions in $\mathcal{O}$. Ergo, the complexity of Algorithm 1 is $O(\mathcal{O}^2 n)$.

---

**Algorithm 1**. The Interval Outranking-based Ant Colony Optimization algorithm

**Input**: $\lambda$, $\vec{w}$, $\vec{v}$, $\vec{q}$, $\varsigma$ and $\xi$
**Output**: An approximation of the best compromise solution ($\mathcal{F}_1$)
1. Randomly initialize the pheromone trail ($\tau$)
2. `Normalize`($\tau$)
3. Rank solutions in $\tau$
4. $t \leftarrow 0$
5. **while** $t < iter_{max}$ **do**
6.     **for each** ant $a \in A$
7.         Generate a solution based on $\tau$
8.     **end for**
9.     $\mathcal{O} \leftarrow A \cup \tau$
10.     `Normalize`($\mathcal{O}$)
11.     Rank solutions in $\mathcal{O}$

12.   $\tau \leftarrow \emptyset$
13.   Copy into $\tau$ the first $\kappa$ elements of $\mathcal{O}$
14.   $t \leftarrow t + 1$
**15. end while**
16. **return** $\tau$

## 5. Experimental results

We implemented IO-ACO using C under Linux (Ubuntu 18) on a computer with an Intel Core i7-6700 3.4 GHz with 16GB of RAM. All experiments reported here were conducted in that computer setting. The parameter values of IO-ACO are $\varsigma=0.1$ and $\xi=0.5$ (cf. Falcon-Cardona & Coello Coello, 2017).

For each problem tested in this section, IO-ACO was run 300 times; it considers DM preferences; consequently, 10 outranking parameter settings representing different DMs were run 30 times. The reference algorithms (iMOACO$_R$ and MOEA/D) were also run 300 times. Hereon, the adjective 'significant' means a Wilcoxon's non-parametric statistical test with a 0.95-confidence level.

*5.1 Benchmark problems*

DTLZ (Deb et al., 2002) and WFG (Huband et al., 2005) have become the standard test suites used to assess the performance of algorithms to solve problems with multiple objective functions. These continuous problems are scalable with respect to the number of objective functions and decision variables; additionally, they offer Pareto frontiers with a wide range of properties (i.e., concavity, convexity, multi-frontality, linearity, bias, connectivity, degeneration, and separability).

In this paper, we have run IO-ACO on the nine problems for both the DTLZ and the WFG test suites, named DTLZ1–DTLZ9 and WFG1–WFG9. Also, we explored each problem varying the number of objectives, considering 3, 5, 7 and 10 objective functions. Table 1 summarizes the settings for the standard problems, including the number of decision variables and the position-related variables.

Table 1. Parameters for the Standard Problems used in the experiments

| Problems | Number of objectives ($m$) | Position ($k$) | Number of decision variables ($n$) |
|---|---|---|---|
| DTLZ1 | {3, 5, 7, 10} | 5 | $m + k - 1$ |
| DTLZ2–DTLZ6 | {3, 5, 7, 10} | 10 | $m + k - 1$ |
| DTLZ7 | {3, 5, 7, 10} | 20 | $m + k - 1$ |
| DTLZ8, DTLZ9 | {3, 5, 7, 10} | $m-1$ | $10m$ |
| WFG1–WFG9 | 3 | $2(m-1)$ | 24 |
| WFG1–WFG9 | 5 | $2(m-1)$ | 47 |
| WFG1–WFG9 | 7 | $2(m-1)$ | 70 |
| WFG1–WFG9 | 10 | $2(m-1)$ | 105 |

.

*5.2 Performance assessment*

Unlike most multiobjective metaheuristics, IO-ACO is not intended to approximate uniformly distributed samples of the Pareto frontier. Hence, most of the current metrics are not adequate to assess its performance (e.g., spread, spacing, and hypervolume). IO-ACO searches for the solutions that meet the conditions to be the best compromise solution, according to the relational system of fuzzy preferences. Thus, the Region of Interest (RoI) is the set of solutions satisfying Problem 3. Because the RoI is a subset of the Pareto frontier, IO-ACO would be competitive if it generates solutions that are close enough to the true RoI. With this aim in mind, we approximated a 100000-point sample of the Pareto frontier for each problem and applied the outranking model of Section 3.2 to identify the RoI. The distance to this approximated RoI (named A-RoI hereon) is used to measure the quality of the solutions provided by IO-ACO in every single run; particularly, we consider the Euclidean distance and the Chebyshev distance. In line with this notion, given the state of the last set of solutions $X$ of an algorithm, denoted $X^*$, and the A-RoI of a particular DM for a specific problem, the following four indicators are utilized:

- Minimum Euclidian distance. This indicator is the Euclidean distance between the closest solution from $X^*$ to the A-RoI.
- Average Euclidian distance. It is the average Euclidean distance among the solutions from $X^*$ to those of the A-RoI.
- Minimum Chebyshev distance. This indicator is the Chebyshev distance between the closest solution from $X^*$ to the A-RoI.
- Average Chebyshev distance. It is the average Chebyshev distance among the solutions from $X^*$ to those of the A-RoI.

*5.3 Comparison with a state-of-the-art ACO approach*

Table 2 compares the results obtained by IO-ACO and iMOACO$_R$ on the standard problems. Here, the first and second columns identify, respectively, the test suite and the dimensionality of the problems. The third column presents the list of problems (identified by numbers) in which iMOACO$_R$ is significantly outperformed by IO-ACO in terms of any of the four distance indicators (see Section 5.2). In contrast, the fourth column lists the problems in which iMOACO$_R$ significantly outperforms IO-ACO. The fifth column specifies the indicator being considered.

As shown in Table 2, IO-ACO outperforms iMOACO$_R$ in most of the problems in both test suites. Considering 72 problems —18 standard problems (DTLZ1–9, and WFG1–9) with four levels of objectives (3, 5, 7, and 10)— IO-ACO obtained better results in:
- 53 problems considering the minimum Chebyshev distance,
- 48 problems considering the average Chebyshev distance,
- 51 problems considering the minimum Euclidean distance, and
- 51 problems considering the average Euclidean distance.

These differences were statistically significant. Regardless of the number of objectives, IO-ACO offers solutions that better match the DM preferences, even assuming an effective *a posteriori* multicriteria analysis on the solutions by iMOACO$_R$. According to the results, IO-ACO is more effective on DTLZ than on WFG when is compared with iMOACO$_R$.

*5.4 Comparison with a state-of-the-art evolutionary approach*

Table 3 compares the results obtained by IO-ACO and MOEA/D on the standard problems. The columns of Table 3 have the same meaning as those of Table 2. The results may be summarized as follows:
- Considering the minimum Chebyshev distance, IO-ACO obtained better results in 38 problems and

MOEA/D in 28 (there are six problems without a statistically significant difference). IO-ACO is more effective on the WFG test suite than on DTLZ when is compared with respect to MOEA/D.
- Considering the average Chebyshev distance, IO-ACO obtained better results in 44 problems and MOEA/D in 26 (there are two problems without a statistically significant difference). Compared with MOEA/D, IO-ACO is especially effective when solving the DTLZ test problems with 10 objectives.
- Considering the minimum Euclidean distance, IO-ACO obtained better results in 42 problems and MOEA/D in 27 (there are three problems without a statistically significant difference). The advantages of IO-ACO over MOEA/D are more evident in the WFG test problems with 10 objectives; additionally, in WFG, the performance of IO-ACO increased as the number of objectives scaled.

Table 2. Comparison between IO-ACO and iMOACO$_R$

| Benchmark | Number of objectives | Problems in which iMOACO$_R$ (a) is outperformed by O-ACO | (b) outperforms O-ACO | Indicator |
|---|---|---|---|---|
| DTLZ | 3 | 1, 3, 4, 5, 6, 8, 9 | 7 | Min. Chevishev |
|  | 5 | 1, 2, 3, 4, 5, 6, 8 | 9 |  |
|  | 7 | 1, 2, 3, 4, 5, 6, 8 | 9 |  |
|  | 10 | 1, 2, 3, 4, 5, 6, 7, 9 |  |  |
| WFG | 3 | 1, 3, 4, 6, 7, 8, 9 | 2 |  |
|  | 5 | 1, 2, 3, 7, 9 | 5, 6, 8 |  |
|  | 7 | 1, 2, 4, 5, 8, 9 | 6, 7 |  |
|  | 10 | 1, 3, 5, 6, 7, 9 | 2, 8 |  |
| Counting of problems |  | 53 | 11 |  |
| DTLZ | 3 | 1, 3, 4, 5, 8, 9 | 2, 7 | Avg. Chevishev |
|  | 5 | 1, 3, 5, 6, 8 | 4, 9 |  |
|  | 7 | 1, 2, 3, 5, 6, 8 |  |  |
|  | 10 | 1, 3, 5, 6, 7, 9 | 4 |  |
| WFG | 3 | 1, 3, 4, 6, 7, 8 | 2, 5, 9 |  |
|  | 5 | 1, 3, 4, 5, 9 | 2, 6, 7, 8 |  |
|  | 7 | 1, 2, 3, 4, 5, 7, 8, 9 |  |  |
|  | 10 | 1, 3, 4, 6, 7, 8, 9 | 2, 5 |  |
| Counting of problems |  | 48 | 14 |  |
| DTLZ | 3 | 1, 3, 4, 5, 6, 7, 8, 9 | 2 | Min. Euclidean |
|  | 5 | 1, 2, 3, 4, 5, 6, 7 | 8 |  |
|  | 7 | 1, 2, 3, 4, 5, 6, 7, 8 | 9 |  |
|  | 10 | 1, 2, 3, 4, 5, 6, 9 | 7 |  |
| WFG | 3 | 1, 3, 4, 6, 8, 9 | 2, 5 |  |
|  | 5 | 2, 4, 5, 9 | 6, 7, 8 |  |
|  | 7 | 2, 3, 4, 8, 9 | 5, 6, 7 |  |
|  | 10 | 1, 3, 5, 6, 7, 9 | 2, 8 |  |
| Counting of problems |  | 51 | 14 |  |
| DTLZ | 3 | 1, 3, 4, 5, 6, 7, 8, 9 |  | Avg. Euclidean |
|  | 5 | 1, 2, 3, 4, 5, 6, 7 | 8, 9 |  |
|  | 7 | 1, 2, 3, 4, 5, 6, 7, 8 |  |  |
|  | 10 | 1, 2, 3, 4, 5, 6, 9 |  |  |

| Benchmark | Number of objectives | Problems in which MOEA/D (a) is outperformed by IO-ACO | (b) outperforms IO-ACO | Indicator |
|---|---|---|---|---|
| WFG | 3 | 1, 4, 6, 7, 8 | 2, 3, 9 | |
| | 5 | 1, 2, 4, 7, 9 | 5, 6, 8 | |
| | 7 | 2, 3, 4, 7, 8, 9 | 5 | |
| | 10 | 1, 3, 6, 7, 9 | 4, 5, 8 | |
| Counting of problems | | 51 | 12 | |

- Considering the average Euclidean distance, IO-ACO obtained better results in 44 problems and MOEA/D in 23 (there are five problems without a statistically significant difference). IO-ACO is more effective on the DTLZ test suite than on WFG when is compared with respect to MOEA/D, and the best relative performance was reached on the DTLZ test problems with five objective functions.

Table 3. Comparison between IO-ACO and MOEA/D

| Benchmark | Number of objectives | Problems in which MOEA/D (a) is outperformed by IO-ACO | (b) outperforms IO-ACO | Indicator |
|---|---|---|---|---|
| DTLZ | 3 | 4, 5, 6, 9 | 1, 3, 7, 8 | Min. Chevishev |
| | 5 | 2, 6, 8, 9 | 1, 3, 5, 7 | |
| | 7 | 2, 4, 7, 9 | 1, 3, 5, 8 | |
| | 10 | 2, 4, 7, 9 | 1, 3, 5 | |
| WFG | 3 | 2, 3, 6, 7, 9 | 1, 4, 5 | |
| | 5 | 1, 2, 3, 4, 6, 7 | 5, 8, 9 | |
| | 7 | 1, 2, 3, 4, 6, 8 | 5, 7, 9 | |
| | 10 | 1, 3, 4, 7, 9 | 2, 5, 6, 8 | |
| Counting of problems | | 38 | 28 | |
| DTLZ | 3 | 2, 4, 5, 6, 9 | 1, 3, 7, 8 | Avg. Chevishev |
| | 5 | 1, 2, 3, 4, 6, 9 | 5, 7, 8 | |
| | 7 | 1, 2, 4, 5, 7, 9 | 3, 6, 8 | |
| | 10 | 2, 4, 5, 7, 8, 9 | 1, 3 | |
| WFG | 3 | 2, 3, 5, 6, 7 | 1, 4, 9 | |
| | 5 | 1, 2, 3, 4, 6, 7 | 5, 8, 9 | |
| | 7 | 1, 3, 4, 6, 8 | 2, 5, 7, 9 | |
| | 10 | 1, 3, 4, 8, 9 | 2, 5, 6, 7 | |
| Counting of problems | | 44 | 26 | |
| DTLZ | 3 | 4, 6, 7, 8, 9 | 1, 2, 3 | Min. Euclidean |
| | 5 | 2, 4, 6, 7, 9 | 1, 3, 5, 8 | |
| | 7 | 2, 4, 7, 9 | 1, 3, 5, 6, 8 | |
| | 10 | 2, 4, 7, 9 | 1, 3, 5, 6 | |
| WFG | 3 | 2, 3, 6, 7, 9 | 1, 4, 5, 8 | |
| | 5 | 1, 2, 4, 6, 7 | 3, 8, 9 | |
| | 7 | 1, 2, 3, 4, 6, 8 | 5, 7, 9 | |
| | 10 | 1, 2, 3, 4, 5, 7, 8, 9 | 6 | |
| Counting of problems | | 42 | 27 | |
| DTLZ | 3 | 2, 4, 5, 6, 7, 8, 9 | 1, 3 | Avg. Euclidean |
| | 5 | 1, 2, 3, 4, 6, 7, 9 | | |
| | 7 | 1, 2, 4, 5, 7, 9 | 3, 6, 8 | |

|       |    |                  |            |
|-------|----|------------------|------------|
|       | 10 | 2, 4, 5, 7, 8, 9 | 1, 3, 6    |
| WFG   | 3  | 2, 5, 6, 7       | 1, 4, 8, 9 |
|       | 5  | 1, 3, 4, 7       | 5, 6, 8, 9 |
|       | 7  | 1, 2, 3, 4, 6, 8 | 5, 7, 9    |
|       | 10 | 1, 2, 3, 9       | 4, 5, 6, 7 |
| Counting of problems | | 44 | 23 |

According to Table 3, IO-ACO provides better solutions than MOEA/D on a regular basis. The effectiveness of our approach depends on the problem, the distance indicator, and the number of objective functions. In spite of this fact, we observed the following consistent patterns for many-objective problems regardless of the selected indicator:

- With seven objectives functions, IO-ACO systematically outperformed MOEA/D in DTLZ2, DTLZ4, DTLZ7, DTLZ9, WFG1, WFG3, WFG4, WFG6, and WFG8.
- With ten objectives functions, IO-ACO systematically outperformed MOEA/D in DTLZ2, DTLZ4, DTLZ7, DTLZ9, WFG1, WFG3, and WFG9.

DTLZ2 and DTLZ4 are multi-frontal; additionally, these problems are Pareto many-to-one. Their objectives are non-separable, and the geometry of the Pareto frontier is concave. Remarkably, the Pareto optimal front of DTLZ4 is biased. DTLZ7 is singularly challenging because the Pareto frontier is disconnected and has mixed concave/convex regions, and the fitness landscape is one-to-one. Unlike the aforementioned DTLZ problems, DTLZ9 has side constraints; its Pareto front is many-to-one and partially degenerate (Huband et al., 2006).

On the other hand, the WFG suite is also challenging because these problems are many-to-one and have no extremal nor medial parameters. Regarding geometry, the Pareto front of WFG1 is convex; WFG3 has a Pareto frontier that is linear and degenerate; and the Pareto front is concave for WFG4, WFG6, WFG8 (which is also biased) and WFG9 (which is also biased).

*5.5 Comparison among IO-ACO, MOEA/D, and iMOACO$_R$*

In this section, we compare the effectiveness of the three approaches on both test suites simultaneously. For each of the 72 problems, the algorithms are sorted according to the conducted statistical tests and post-hoc Holm-Bonferroni analysis. So, for each problem, the best algorithm obtains position 1, and the worst one gets position 3 (in case of a draw, the position would be averaged). Then, the Borda score is calculated to rank the algorithms based on the ranking over every single problem; consequently, the Borda sum would provide a general ranking of the algorithms according to their average performance.

Table 4 presents the Borda score of IO-ACO, MOEA/D, and iMOACO$_R$ on the standard benchmarks. Accordingly, the results allow us to draft the following conclusions:

- IO-ACO showed a better capacity to approach the RoI than MOEA/D and iMOACO$_R$ regardless of the distance indicator and the number of objectives.
- IO-ACO performed better when Euclidean distance-based indicators are considered. This behavior is expected because MOEA/D updates solutions based on Chebyshev distances.
- IO-ACO performed especially well in problems with ten objective functions when the minimum Euclidean distance is considered.
- Let's think of IO-ACO as an extension of iMOACOR. These results are clear evidence of the impact on

the performance when DM preferences are incorporated, increasing the capability to reach the RoI. iMOACO$_R$ was only able to outperform MOEA/D when enriched with the interval outranking system of preferences, originating our version named IO-ACO.

## 6. Conclusions and directions for future research

This paper has proposed a metaheuristic approach—named IO-ACO—to solve many-objective optimization problems through an ACO algorithm enriched with the preferences of the DM articulated in an interval outranking-based system of preferences. This preference model was considered in the solution sorting of the pheromone trail to bias the search towards the best compromise. One of the motivations was that interval outranking is robust enough to model imprecise preferences. As far as we know, IO-ACO is the first ant colony-based metaheuristic using the interval outranking relations to get the edge in solving problems with many objectives to optimize.

Table 4. Comparison among IO-ACO, MOEA/D, and iMOACO$_R$

| Indicator | Number of Objectives | The Borda score of (a) IO-ACO | (b) MOEA/D | (c) iMOACO$_R$ |
|---|---|---|---|---|
| Min. Chevishev | 3 | 30.5 | 34.5 | 43.0 |
| | 5 | 30.0 | 37.0 | 41.0 |
| | 7 | 28.0 | 37.0 | 43.0 |
| | 10 | 31.5 | 34.5 | 42.0 |
| Sum | | 120.0 | 143.0 | 169.0 |
| Avg. Chevishev | 3 | 32.0 | 35.0 | 41.0 |
| | 5 | 29.5 | 44.5 | 34.0 |
| | 7 | 29.5 | 35.0 | 43.5 |
| | 10 | 31.0 | 38.0 | 39.0 |
| Sum | | 122.0 | 152.5 | 157.5 |
| Min. Euclidean | 3 | 28.0 | 38.5 | 41.5 |
| | 5 | 31.5 | 37.5 | 39.5 |
| | 7 | 32.0 | 38.0 | 38.0 |
| | 10 | 26.5 | 41.0 | 40.5 |
| Sum | | 118.0 | 155.0 | 159.5 |
| Avg. Euclidean | 3 | 29.5 | 37.5 | 41.0 |
| | 5 | 27.0 | 40.5 | 40.5 |
| | 7 | 28.0 | 38.0 | 42.0 |
| | 10 | 30.5 | 37.0 | 40.5 |
| Sum | | 115.0 | 153.0 | 164.0 |

IO-ACO convergence to the Region of Interest (RoI) was tested by measuring the distance from the Approximated RoI (A-RoI) to the solution set generated by IO-ACO; four indicators—based on the Euclidean distance and the Chebyshev distance—were used to determine closeness.

The quality of the solutions by IO-ACO was validated through statistically meaningful comparisons with two competitive metaheuristic algorithms, i.e., iMOACO$_R$ and MOEA/D. Regarding iMOACO$_R$, IO-ACO offered solutions closer to the RoI, especially in problems from the DTLZ test suite. Regarding MOEA/D, IO-ACO also provided better solutions (in terms of outranking); however, the advantages are more significant in the WFG test suite.

Overall, we suggest using IO-ACO under the presence of many objective functions because its best performance was reached in problems with ten objectives when the Euclidean indicators were taken. Additionally, IO-ACO was particularly competitive in problems with a Pareto front whose geometry is considered challenging; we mean frontiers that are disconnected and have mixed concave/convex regions (e.g., DTLZ7).

Although it is true that outranking approaches require close interaction with the DM to reach a setting that acceptably reflects their preferences, this paper presented evidence that such an effort may be advantageously compensated in the framework of many-objective optimization via swarm-intelligence metaheuristics.

Further research is needed to provide explanations connecting the performance, the properties of the problem, and the number of objective functions.

**Acknowledgements**

Carlos A. Coello Coello acknowledges support from CONACyT grant no. 1920 and from SEP-Cinvestav grant (application no. 4).

**References**

Abouhawwash, M., & Deb, K. (2021). Reference point based evolutionary multi-objective optimization algorithms with convergence properties using KKTPM and ASF metrics. Journal of Heuristics, 1–40. https://doi.org/10.1007/s10732-021-09470-4

Acciarini, G., Izzo, D., & Mooij, E. (2020). MHACO: A Multiobjective Hypervolume-Based Ant Colony Optimizer for Space Trajectory Optimization. In 2020 IEEE Congress on Evolutionary Computation (CEC) (pp. 1–8). IEEE. https://doi.org/10.1109/CEC48606.2020.9185694

Balderas, F., Fernandez, E., Gómez, C., Rivera, G., Cruz-Reyes, L., & Rangel-Valdez, N. (2016). Uncertainty modelling for project portfolio problem using interval analysis, *International Journal of Combinatorial Optimization Problems and Informatics* 7 (3) (2016) 20–27. https://ijcopi.org/ojs/article/view/24

Balderas, F., Fernandez, E., Gomez, C., Rangel, N., Cruz-Reyes, L. (2019). An interval-based approach for evolutionary multi-objective optimisation of project portfolios. International Journal of Information Technology & Decision Making 18 (4), 1317–1358. https://doi.org/10.1142/S021962201950024X

Bastiani, S., Cruz-Reyes, L., Fernandez, E., & Gomez, C. (2015). Portfolio optimization from a set of preference ordered projects using an ant colony based multiobjective approach, *International Journal of Computational Intelligence Systems* 41–53. https://doi.org/10.1080/18756891.2015.1129590

Bechikh S., Elarbi M., Ben Said L., (2017). Many-objective Optimization Using Evolutionary Algorithms: A Survey, in: Recent Adv. Evol. Multi-Objective Optim., Springer, Cham, 2017, pp. 105–137. https://doi.org/10.1007/978-3-319-42978-6_4.

Branke, J., Kaußler, T., Schmeck, H. (2001) Guidance in evolutionary multi-objective optimization, Adv. Eng. Softw. (32) 499–507. https://doi.org/10.1016/S0965-9978(00)00110-1.

Branke J., Deb K., (2005). Integrating user preferences into evolutionary multiobjective optimization, Tech. Rep. KanGAL 2005, Indian Institute of Technology.

Branke, J., Corrente, S., Greco, S., Słowinski, R. and Zielniewicz, P. (2016). Using Choquet integral as preference model in interactive evolutionary multiobjective optimization, European Journal of Operational Research, 250(3), 884–901. https://doi.org/10.1016/j.ejor.2015.10.027.

Brans, J., & Mareschal, B. (2005). Multiple criteria decision analysis: state of the art surveys, International Series on Operations Research&Management Science, Springer–Verlag, Berlin, 2005, Ch. Promethee methods, pp. 163–190. https://doi.org/10.1007/978-1-4939-3094-4_6

Brockhoff, D.,Wagner, T.,&Trautmann, H. (2012). On the properties of the R2 indicator. In 2012 Genetic and evolutionary computation conference (GECCO'2012) (pp. 465–472). Philadelphia: ACM Press. ISBN: 978-1-4503-1177-9. https://doi.org/10.1145/2330163.2330230

Brockhoff D., Bader J., Thiele L., Zitzler E. (2013). Directed multiobjective optimization based on the weighted hypervolume indicator, Journal of Multi-Criteria Decision Analysis 20 (5-6) 291–317. https://doi.org/10.1002/mcda.1502

Chandra Mohan B., Baskaran R., (2012). A survey: Ant Colony Optimization based recent research and implementation on several engineering domain,Expert Systems with Applications,Volume 39, Issue 4, 2012, Pages 4618–4627, ISSN 0957-4174, https://doi.org/10.1016/j.eswa.2011.09.076.

Chica M., Cordon O., Damas S. and Bautista J., (2009). Integration of an EMO-based Preference Elicitation Scheme into a Multi-objective ACO Algorithm or Time and Space Assembly Line Balancing, 2009 IEEE Symposium on Computational Intelligence in Multi-Criteria Decision-Making (MCDM'2009), IEEE Press, Nashville, TN, SA, pp.157–162, 2009, ISBN 978-1-4244-2764-2. https://doi.org/10.1109/MCDM.2009.4938843

.Chica, M., Cordón, O., Damas, S., & Bautista, J. (2010). Multi-objective, constructive heuristics for the 1/3 variant of the time and space assembly line balancing problem: ACO and random greedy search. Information Sciences, 180, 3465–3487. https://doi.org/10.1016/j.ins.2010.05.033

Chica M., Cordon O., Damas S. and Bautista J., (2011). Including different kinds of preferences in a multi-objective ant algorithm for time and space assembly line balancing on different Nissan scenarios, Expert Systems with Applications 2011, Volumen 1, pp.709–720. https://doi.org/10.1016/j.eswa.2010.07.023

Cruz-Reyes, L., Fernandez, E., Gomez, C., Rivera, G., & Perez, F. (2014). Many-objective portfolio optimization of interdependent projects with 'a priori' incorporation of decision-maker preferences, *Applied Mathematics & Information Sciences* 8 (4) 1517. https://doi.org/10.12785/amis/080405

Cruz-Reyes, L., Fernandez, E., Sanchez, P., Coello Coello, C.A. and Gomez, C., (2017). Incorporation of implicit decision-maker preferences in multi-objective evolutionary optimization using a multi-criteria classification method, Appl. Soft Comput. J., (50), 48–57. https://doi.org/10.1016/j.asoc.2016.10.037.

Cruz-Reyes, L., Fernandez, E., Sanchez-Solis, J.P., Coello Coello, C.A. and Gomez, C., (2020). Hybrid evolutionary multi-objective optimisation using outranking-based ordinal classification methods, Swarm and Evolutionary Computation, 54 (100652), https://doi.org/10.1016/j.swevo.2020.100652.

Cvetkovic, D., Parmee, I.C. (2002) Preferences and their application in evolutionary Multiobjective optimization, IEEE Trans. Evol. Comput. 6, 42–57. https://doi.org/10.1109/4235.985691.

Deb, K., Thiele, L., Laumanns, M., &Zitzler, E. (2002). Scalable multiobjective optimization test problems. In Congress on evolutionary computation (CEC'2002) (vol. 1, pp. 825–830). Piscataway, NJ: IEEE Service Center. https://doi.org/10.1109/CEC.2002.1007032

Deb, K., Sinha, A., Korhonen, P.J. and Wallenius, J. (2010). An interactive evolutionary multiobjective optimization method based on progressively approximated value functions, IEEE Trans. Evol. Comput. 14(5) 723–739. https://doi.org/10.1109/TEVC.2010.2064323.

Dorigo, M. and Stützle, T., (2004). Ant colony optimization. Cambridge, MA: MIT Press.

Dorigo M., Blumb C., Ant colony optimization theory: A survey, Theoretical Computer Science 344 (2005) 243–278. https://doi.org/10.1016/j.tcs.2005.05.020

Du B., Chen H., Huang G.Q., Yang H.D. (2011) Preference Vector Ant Colony System for Minimising Make-span and Energy Consumption in a Hybrid Flow Shop. In: Wang L., Ng A., Deb K. (eds) Multi-objective Evolutionary Optimisation for Product Design and Manufacturing. Springer, London. https://doi.org/10.1007/978-0-85729-652-8_9.

Dunwei Gong, Fenglin Sun, Jing Sun and Xiaoyan Sun, (2017). Set-Based Many-Objective Optimization Guided by a Preferred Region,

Neurocomputing, 228, 241–255. https://doi.org/10.1016/j.neucom.2016.09.081

Falcón-Cardona J.G., Coello Coello C.A. (2016) iMOACO$_\mathbb{R}$: A New Indicator-Based Multi-objective Ant Colony Optimization Algorithm for Continuous Search Spaces. In: Handl J., Hart E., Lewis P., López-Ibáñez M., Ochoa G., Paechter B. (eds) Parallel Problem Solving from Nature – PPSN XIV. PPSN 2016. Lecture Notes in Computer Science, vol 9921. Springer, Cham. https://doi.org/10.1007/978-3-319-45823-6_36

Falcón-Cardona, J.G., Coello Coello, C.A. (2017) A new indicator-based many-objective ant colony optimizer for continuous search spaces. Swarm Intelligence, 11(1), 71–100. https://doi.org/10.1007/s11721-017-0133-x

Fernandez, E., Lopez, E., Bernal, S., Coello Coello, C. A. C., & Navarro, J. (2010). Evolutionary multiobjective optimization using an outranking-based dominance generalization. Computers & Operations Research, 37(2), 390–395. https://doi.org/10.1016/j.cor.2009.06.004

Fernandez, E., Lopez, E., Lopez, F., & Coello Coello, C. A. C. (2011). Increasing selective pressure towards the best compromise in evolutionary multiobjective optimization: The extended NOSGA method. Information Sciences, 181(1), 44–56. https://doi.org/10.1016/j.ins.2010.09.007

Fernandez E., Gomez C., Rivera G., Cruz-Reyes L., Hybrid metaheuristic approach for handling many objectives and decisions on partial support in project portfolio optimisation, Information Sciences, Volume 315, 2015, Pages 102-122, ISSN 0020-0255, https://doi.org/10.1016/j.ins.2015.03.064

Fernandez, Eduardo & Santillán, Claudia & Cruz Reyes, Laura & Rangel-Valdez, Nelson & Bastiani, Shulamith. (2017). Design and Solution of a Surrogate Model for Portfolio Optimization Based on Project Ranking. Scientific Programming. 2017. 1–10. https://doi.org/10.1155/2017/1083062

Fernández E., Figueira J.R., Navarro J., (2019). An indirect elicitation method for the parameters of the ELECTRE TRI-nB model using genetic algorithms, Appl. Soft Comput. 77, 723–733. https://doi.org/10.1016/J.ASOC.2019.01.050.

Fernandez Eduardo, Navarro Jorge, Solares Efrain, Coello Coello Carlos, (2020). Using evolutionary computation to infer the decision maker's preference model in presence of imperfect knowledge: A case study in portfolio optimization, Swarm and Evolutionary Computation, 54 (100648). https://doi.org/10.1016/j.swevo.2020.100648.

Ferreira, T. do N., Araújo, A. A., Basílio Neto, A. D., & de Souza, J. T. (2016). Incorporating user preferences in ant colony optimization for the next release problem. Applied Soft Computing, 49, 1283–1296. doi:10.1016/j.asoc.2016.06.027

Figueira, J., & Roy, B. (2002). Determining the weights of criteria in the ELECTRE type methods with a revised Simos' procedure. European journal of operational research, 139(2), 317–326. https://doi.org/10.1016/S0377-2217(01)00370-8

Fliedner, T., & Liesio, J. (2016). Adjustable robustness for multi-attribute project portfolio selection, *European Journal of Operational Research* 252(3), 931–946., https://doi.org/10.1016/j.ejor.2016.01.058

French, S., (1993). Decision theory: an introduction to the mathematics of rationality, Ellis Horwood, https://books.google.com.mx/books?id=HcxBPwAACAAJ&dq=%22Decision+Theory.+An+introduction+to+the+mathematics+of+rationality%22&hl=es&sa=X&ved=0ahUKEwiX8rPBz8niAhVOIqwKHaE8CKEQ6AEIKTAA (accessed June 1, 2019).

Gomez, C., Cruz-Reyes, L., Rivera, G., Rangel-Valdez, N., Morales-Rodriguez, L., & Perez-Villafuerte, M. (2018) Interdependent projects selection with preference incorporation, in: New Perspectives on Applied Industrial Tools and Techniques, Springer, 2018, pp. 253–271 https://doi.org/10.1007/978-3-319-56871-3_13

Grassé P.-P., "La reconstruction du nid et les coordinations interindividuelles chez bellicoitermes natalenis et cubitermes sp. la théorie de la stigmergie: Essai d'interprétation du comportement des termites constructeurs", Insect Sociaux, vol. 6, pp. 41–81, 1959. https://doi.org/10.1007/BF02223791

He, Y., He, Z., Kim, K. J., Jeong, I. J., & Lee, D. H. (2020). A Robust Interactive Desirability Function Approach for Multiple Response Optimization Considering Model Uncertainty. IEEE Transactions on Reliability. https://doi.org/10.1109/TR.2020.2995752

Helson Luiz Jakubovski Filho, Nascimento Ferreira Thiago and Regina Vergilio Silvia, (2018). Incorporating User Preferences in a Software Product Line Testing Hyper-Heuristic Approach, 2018 IEEE Congress on Evolutionary Computation (CEC'2018),2283–2290,IEEE Press, Rio de Janeiro, Brazil, July 8-13, ISBN: 978-1-5090-6017-7. https://doi.org/10.1109/CEC.2018.8477803

Hernández Gómez, R., & Coello Coello, C. A. (2015). Improved metaheuristic based on the R2 indicator for many-objective optimization. In 2015 Genetic and evolutionary computation conference (GECCO 2015) (pp. 679–686). Madrid: ACM Press. ISBN 978-1-4503-3472-3. https://doi.org/10.1145/2739480.2754776

Huband, S., Barone, L., While, L., & Hingston, P. (2005). A scalable multiobjective test problem toolkit. In C. A. Coello Coello, A. Hernández Aguirre, & E. Zitzler (Eds.), Third international conference evolutionary multi-criterion optimization (EMO 2005), lecture notes in computer science (vol. 3410, pp. 280–295). Springer. https://doi.org/10.1007/978-3-540-31880-4_20

Huband, S., Hingston, P., Barone, L., & While, L. (2006). A review of multiobjective test problems and a scalable test problem toolkit. *IEEE Transactions on Evolutionary Computation*, 10(5), 477–506. https://doi.org/10.1109/TEVC.2005.861417

Ishibuchi, H., Matsumoto, T., Masuyama, N., & Nojima, Y. (2020). Effects of dominance resistant solutions on the performance of evolutionary multi-objective and many-objective algorithms. In Proceedings of the 2020 Genetic and Evolutionary Computation Conference (pp. 507–515). https://doi.org/10.1145/3377930.3390166

Kulturel-Konak, S., Coit, D. W., & Baheranwala, F. (2008). Pruned Pareto-optimal sets for the system redundancy allocation problem based on multiple prioritized objectives. Journal of Heuristics, 14(4), 335. https://doi.org/10.1007/s10732-007-9041-3

Li, K. (2019). Progressive preference learning: Proof-of-principle results in MOEA/D. In International Conference on Evolutionary Multi-Criterion Optimization, Springer, Cham. 631–643. https://doi.org/10.1007/978-3-030-12598-1_50

Li, K., Liao, M., Deb, K., Min, G., Yao, X.: Does preference always help? A holistic study on preferencebased evolutionary multi-objective optimisation using reference points. IEEE Trans. Evolut. Comput. 24(6), 1078–1096. https://doi.org/10.1109/TEVC.2020.2987559

Liu, R., Zhou, R., Ren, R., Liu, J., Jiao, L.: Multi-layer interaction preference based multi-objective evolutionary algorithm through decomposition. Inf. Sci. 509, 420–436 (2020). https://doi.org/10.1016/j.ins.2018.09.069

Lücken V, Brizuela C., & Barán, B. (2019). An overview on evolutionary algorithms for many-objective optimization problems. Wiley interdisciplinary reviews: data mining and knowledge discovery, 9(1), e1267. https://doi.org/10.1002/widm.1267

Mandal A.K., Dehuri S. (2020) A Survey on Ant Colony Optimization for Solving Some of the Selected NP-Hard Problem. In: Dehuri S., Mishra B., Mallick P., Cho SB., Favorskaya M. (eds) Biologically Inspired Techniques in Many-Criteria Decision Making. BITMDM 2019. Learning and Analytics in Intelligent Systems, vol 10. Springer, Cham. https://doi.org/10.1007/978-3-030-39033-4_9

Miettinen K., Ruiz F., and Wierzbicki A. P., "Introduction to multiobjective optimization: Interactive approaches," in Multiobjective Optimization: Interactive and Evolutionary Approaches. Berlin, Germany: Springer, 2008, pp. 27–57

Miller, G.A. (1956). The magical number seven, plus or minus two: some limits on our capacity for processing information, Psychol. Rev. 63 (2), 81–97. https://doi.org/10.1037/h0043158.

Moore, R.E. (1979). Methods and applications of interval analysis, Studies in Applied and Numerical Mathematics, *Society for Industrial and Applied Mathematics*, https://doi.org/10.1137/1.9781611 970906

Parreiras, R. O., and Vasconcelos, J. A., (2007). A multiplicative version of Promethee II applied to multiobjective optimization problems. European Journal of Operational Research, 183(2), 729–740. https://doi.org/10.1016/j.ejor.2006.10.002

Rangel-Valdez, N., Gómez-Santillán, C., Hernández-Marín, J., Morales-Rodriguez, M., Cruz-Reyes, L., & Fraire-Huacuja, H. (2020). Parallel designs for metaheuristics that solve portfolio selection problems using fuzzy outranking relations, *International Journal of Fuzzy Systems* (2020) 1–13. https://doi.org/10.1007/s40815-019-00794-9

Rivera, G., Gómez, G., Cruz, L., García, R., Balderas, F., Fernández, E., & López, F. (2012). Solution to the social portfolio problem by evolutionary algorithms, *International Journal of Combinatorial Optimization Problems and Informatics* 3 (2) 21–30.

Roy, B., & Vanderpooten, D. (1996). The European school of MCDA: Emergence, basic features and current works. *Journal of Multi-Criteria Decision Analysis* 5, 22–38. https://doi.org/10.1002/(SICI)1099-1360(199603)5:1<22::AID-MCDA93>3.0.CO;2-F

Roy B., (1990). Reading in multiple criteria decision aid, Springer-Verlag, Ch. The outranking approach and the foundations of ELECTRE methods, pp. 155–183. https://doi.org/10.1007/978-3-642-75935-2_8

Roy B., (1990). The Outranking Approach and the Foundations of Electre Methods, in: Readings Mult. Criteria Decis. Aid, Springer Berlin Heidelberg, Berlin, Heidelberg, 155–183. https://doi.org/10.1007/978-3-642-75935-2_8.

Saldanha, W., Arrieta, F., Machado-Coelho, T., Santos, E., Maia, C., Ekel, P., & Soares, G. (2019). Evolutionary algorithms and the Preference Ranking Organization Method for Enrichment Evaluations as applied to a multiobjective design of shell-and-tube heat exchangers. Case Studies in Thermal Engineering. 17. 100564. https://doi.org/10.1016/j.csite.2019.100564

Siegmund Florian, Ng Amos H. C. and Deb Kalyanmoy, (2017). A Comparative Study of Fast Adaptive Preference-Guided Evolutionary Multi-objective Optimization, Evolutionary Multi-Criterion Optimization, 9th International Conference, EMO 2017, In. (eds) Heike Trautmann and Gunter Rudolph and Kathrin Klamroth and Oliver Schutze and Margaret Wiecek and Yaochu Jin and Christian Grimme, 560–574, Springer. Lecture Notes in Computer Science Vol. 10173, Munster, Germany, ISBN 978-3-319-54156-3, 2017. https://doi.org/10.1007/978-3-319-54157-0_38

Socha, K., & Dorigo, M. (2008). Ant colony optimization for continuous domains. *European Journal of Operational Research*, 185(3), 1155–1173. https://doi.org/10.1016/j.ejor.2006.06.046

Taboada HA, Baheranwala F, Coit DW. Practical solutions for multi-objective optimization: an approach to system reliability design problems. Reliability Engineering and System Safety 2007; 92:314–22. https://doi.org/10.1016/j.ress.2006.04.014

Tomczyk, M. K. and Kadziński, M. Decomposition-Based Interactive Evolutionary Algorithm for Multiple Objective Optimization, in IEEE Transactions on Evolutionary Computation, 24(2), 320–334, https://doi.org/10.1109/TEVC.2019.2915767

Wagner T., and Trautmann H., (2010). Integration of preferences in hypervolume-based multiobjective evolutionary algorithms by means of desirability functions, IEEE Trans. Evol. Comput. 14, 688–701. https://doi.org/10.1109/TEVC.2010. 2058119.

Wang R, Purshouse RC, Fleming PJ (2015) Preference-inspired co-evolutionary algorithms using weight vectors. Eur J Oper Res 243(2):423–441. https://doi.org/10.1016/j.ejor.2014.05.019

Xin, B., Chen, L., Chen, J., Ishibuchi, H., Hirota, K., & Liu, B. (2018). Interactive multiobjective optimization: A review of the state-of-the-art. IEEE Access, 6, 41256–41279. https://doi.org/10.1109/ACCESS.2018.2856832

Yali Wang, Steffen Limmer, Markus Olhofer, Michael Emmerich, Thomas Baeck, Automatic Preference Based Multi-objective Evolutionary Algorithm on Vehicle Fleet Maintenance Scheduling Optimization, arXiv preprint arXiv:2101.09556.

Yi Jun, Bai Junren, He Haibo, Peng Jun and Tang Dedong, (2019). ar-MOEA: A Novel Preference-Based Dominance Relation for Evolutionary Multiobjective Optimization, IEEE Transactions on Evolutionary Computation, 23 (5), 788–802. https://doi.org/10.1109/TEVC.2018.2884133

Yuan, M. H., Chiueh, P. T., & Lo, S. L. (2021). Measuring urban food-energy-water nexus sustainability: Finding solutions for cities. Science of The Total Environment, 752, 141954. https://doi.org/10.1016/j.scitotenv.2020.141954

Yutao Qi, Xiaodong Li, Jusheng Yu, Qiguang Miao, (2019). User-preference based decomposition in MOEA/D without using an ideal point, Swarm and Evolutionary Computation, 44, 597–611. https://doi.org/10.1016/j.swevo.2018.08.002.

Zhang, Q., Li, H. MOEA/D: A Multiobjective Evolutionary Algorithm Based on Decomposition. IEEE Transactions on Evolutionary Computation 2007, Volume 11, pp. 712–731. https://doi.org/10.1109/tevc.2007. 892759.

Zhang, Y., Yang, Di Long Guo, Zhen Xu Sun & Da Wei Chen (2019). A novel CACOR-SVR multiobjective optimization approach and its application in aerodynamic shape optimization of high-speed train. *Soft Computing*, 23, 5035–5051. https://doi.org/10.1007/s00500-018-3172-3